\documentclass[conference]{IEEEtran}
\IEEEoverridecommandlockouts
\usepackage{cite}
\usepackage{amsmath,amssymb,amsfonts}
\usepackage{graphicx}
\usepackage{textcomp}
\usepackage{xcolor}
\usepackage{lipsum}
\usepackage{multicol}
\usepackage{subcaption}
\usepackage{soul}
\usepackage{bm}
\usepackage{float}
\usepackage{algorithm,algpseudocode} 
\usepackage{amssymb, amsmath}
\usepackage{graphics}
\usepackage{url}
\usepackage{rotating}
\usepackage[compact]{titlesec}
\usepackage{pseudocode}  
\usepackage{algpascal}
\usepackage{algc} 
\usepackage{adjustbox}
\usepackage{epstopdf}
\usepackage{epsfig}
\usepackage{color}
\usepackage{multirow} 
\usepackage{url}  
\usepackage{caption}
\usepackage{tabularx}
\usepackage{textcomp}
\usepackage{pgfplots}
\pgfplotsset{width=10cm,compat=1.9}

\usepackage{titlesec}

\RequirePackage{enumitem}

\setlist[enumerate]{itemsep=0pt, topsep=0pt, itemindent=15pt, leftmargin=0pt,listparindent=\parindent}
\setlist[itemize]{itemsep=0pt, topsep=0pt, itemindent=15pt, leftmargin=0pt,listparindent=\parindent}

\def\BibTeX{{\rm B\kern-.05em{\sc i\kern-.025em b}\kern-.08em
    T\kern-.1667em\lower.7ex\hbox{E}\kern-.125emX}}
\newcommand{\ignore}[1]{}

\begin{document}
\title{Hardware-Software Co-optimised Fast and Accurate Deep Reconfigurable Spiking Inference Accelerator Architecture Design Methodology}
\author{\IEEEauthorblockN{Anagha Nimbekar$^1$, Prabodh Katti$^2$, Chen Li$^2$, Bashir M. Al-Hashimi$^2$, Amit Acharyya$^1$, Bipin Rajendran$^2$}
\IEEEauthorblockA{
$^1$Indian Institute of Technology, Hyderabad, India; $^2$King's College London, United Kingdom \\
Email: ee20resch01001@iith.ac.in, prabodh.katti@kcl.ac.uk, chen.7.li@kcl.ac.uk, \\ bashir.al-hashimi@kcl.ac.uk, amit\_acharyya@iith.a.c.in, bipin.rajendran@kcl.ac.uk$^{\ast}$}
}

\maketitle

\begin{abstract} 
Spiking Neural Networks (SNNs) have emerged as a promising approach to improve the energy efficiency of machine learning models, as they naturally implement event-driven computations while avoiding expensive multiplication operations.In this paper, we develop a hardware-software co-optimisation strategy to port software-trained deep neural networks (DNN) to reduced-precision spiking models demonstrating fast and accurate inference in a novel event-driven CMOS reconfigurable spiking inference accelerator. Experimental results show that a reduced-precision Resnet-18 and VGG-11 SNN models achieves classification accuracy within 1\% of the baseline full-precision DNN model within 8 spike timesteps. We also demonstrate an FPGA prototype implementation of the spiking inference accelerator with a throughput of $38.4$ giga operations per second (GOPS) consuming $1.54\,$ Watts on PYNQ-Z2 FPGA. This corresponds to $0.6\,$GOPS per processing element and $2.25\,$GOPS/DSP slice,  which is $2\times$ and $4.5\times$ higher utilisation efficiency respectively compared to the state-of-the-art.  Our co-optimisation strategy can be employed to develop deep reduced precision SNN models and port them to resource-efficient event-driven hardware accelerators for edge applications. 


\end{abstract}


\section{Introduction}
Convolutional Neural Networks (CNNs) have achieved human-equivalent accuracies for a wide variety of cognitive tasks sush as language translation \cite{4}, image recognition \cite{1,he2016deep}, and autonomous driving \cite{5}. However, the execution of
CNNs require substantial computational resources, particularly within their convolutional layers, with the number of multiply-accumulate (MAC) computations necessary for the convolution operations exceeding more than  \textgreater90\% of all operations  \cite{Suda}. This imposes significant challenges for the deployment of CNNs for several edge applications that have stringent 
energy and footprint constraints.

Efficient hardware accelerators demand a careful balance and co-optimisation between algorithms, model architecture, and hardware capabilities. At the algorithms level, one promising avenue worth exploring is the realisation of machine learning models using  Spiking Neural Networks (SNNs) that encode and process information using spikes. A `spike' is a  canonical all-or-none binary event, generated by approximating the integrate-and-fire dynamics of neurons in biological brains. As computation and communication are triggered only when a neuron spikes, such models naturally achieve sparsity in model realisation, leading to potentially higher energy efficiencies compared to the conventional artificial neural network (ANN) models used in machine learning. Binary Neural Networks (BNNs) are another category of networks where activations and weights are binary. BNN expressiveness is thus constrained in comparison to real-valued neural networks. Their performance on difficult tasks requiring a high degree of precision may suffer as a result as mentioned in \cite{8678682} \cite{hubara2016binarized} \cite{rastegari2016xnor} \cite{courbariaux2015binaryconnect}.

 {
\begin{figure}[]
  \centering
    \includegraphics[width=0.5\textwidth]{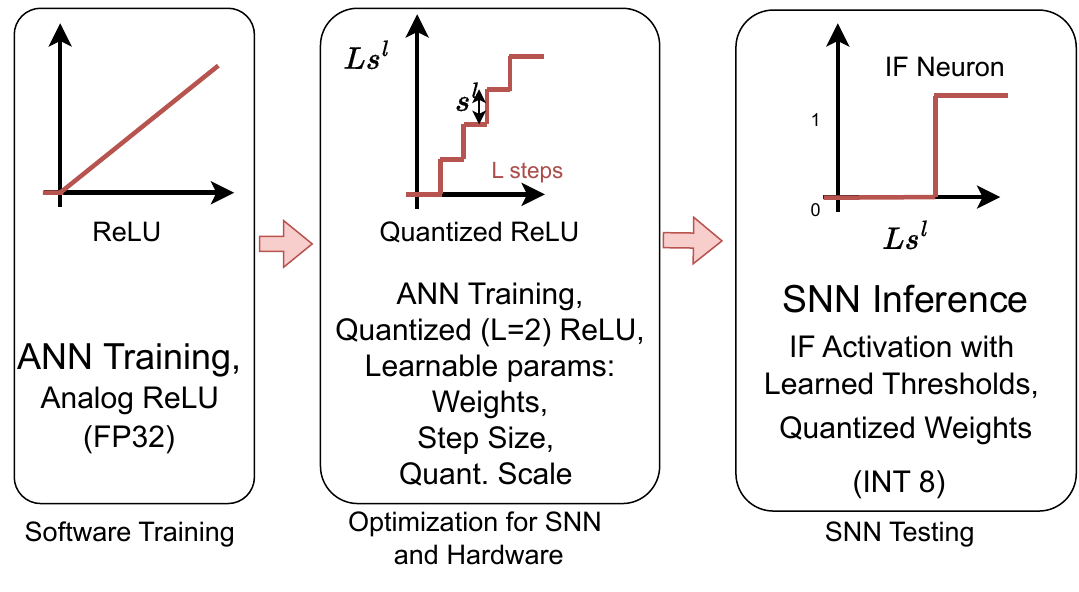}
    \caption{ANN-to-SNN conversion strategy   for developing reduced precision  SNN models for fast inference in hardware.}
  \label{fig2}
\end{figure}}

However, in spite of the significant strides made in developing algorithms for both training SNNs from scratch using surrogate gradient methods \cite{neftci2019surrogate} or by converting  ANNs to SNNs \cite{li2022quantization,bu2023optimal},  it has been observed that most of these networks require hundreds of time steps to match the accuracy of ANNs. Furthermore, it is also crucial that accelerator architectures and algorithmic methods to be co-optimised for efficiency and accuracy, by focusing on reduced precision implementations. In contribution to above study, we employ an ANN-to-SNN conversion approach tailored for low-precision parameter representation, as elaborated in \cite{li2022quantization,bu2023optimal,shymyrbay2022training}.
We also propose a novel hardware-software co-optimized fast and accurate reconfigurable \emph{spiking inference accelerator} (SIA) achieving software-equivalent classification accuracies.
This method remarkably ensures minimal accuracy degradation when realizing SNNs in cutting-edge deep networks such as ResNet-18, and VGG-11. We then demonstrate a highly resource-efficient and reconfigurable hardware architecture that delivers fast inference of such SNN models. 


\begin{figure*}[htbp]
\centering
\captionsetup{justification=centering,margin=0.35cm}
  \includegraphics [scale = 0.5]{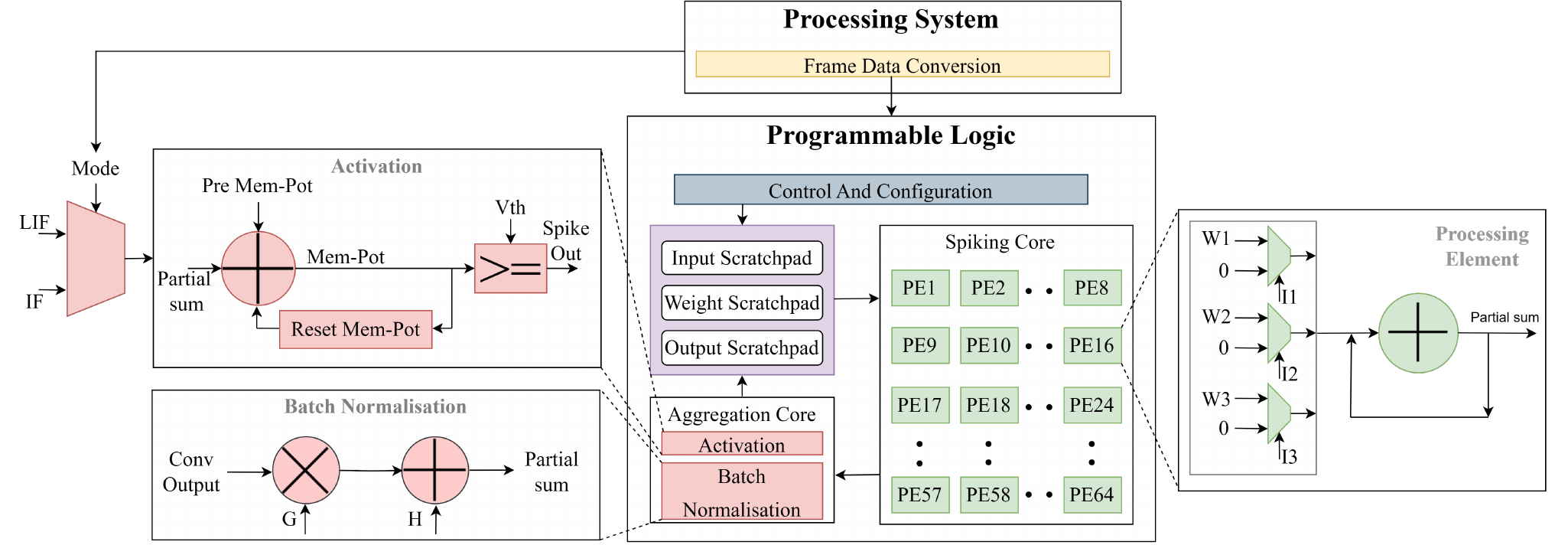}
  \caption{High-level block diagram of the proposed Spiking Inference Accelerator (SIA). The spiking core consists of 64 PE elements to execute spiking convolution while the aggregation core executes neuronal computation. }
  \label{Fig}
\vspace{-2mm}
\end{figure*} 


Algorithms for converting deep  ANN models to SNNs have been reported in \cite{li2022quantization,bu2023optimal}, but these have not been optimised for hardware implementation, use positive and negative spikes and do not consider reduced bit precision network parameters. Quantization of weights for surrogate-gradient trained SNNs have been studied in \cite{shymyrbay2022training}, though this has not been validated for conversion of ANNs to low latency SNNs.
Recent results from literature that focus on hardware acceleration of SNNs can be broadly divided into two classes -- (i) accelerators optimized for fully-connected networks that do not support convolutional layers \cite{ali,nguyen} and (ii) hybrid architecture that has dedicated and separate cores to handle SNN dynamics and convolution operations \cite{tianjic,cdnn}. All these prior works report that SNN accuracy is considerably lower than CNN accuracy, while also typically requiring $>50$ spike timesteps to guarantee near DNN-equivalent performance. 
\cite{gilan2019fpga} proposed an FPGA based accelerator for object detection. Various othervliterateurs (\cite{qiu2016going} \cite{chen2020throughput} \cite{li2021fpga} \cite{guo2017angel}) have shown accelerator architecture for convolutional neural networks to accelerate the inference. However, these architcetures consumes more DSP blocks on FPGA.
\cite{Fang} proposed an FPGA-based SNN implementation with Leaky Integrate and Fire (LIF) neurons for an event-driven MNIST dataset. \cite{Neil} also presented FPGA implementation IF neuron for event-driven data.  
However, it is observed that the above-mentioned literature either has LIF or IF neurons. Also, these literatures works with event-driven data which sets the motivation for this work. Recently, many works have shown improvement in these limitations such as \cite{mayr2019spinnaker}\cite{loihi}.
In this paper, we present an efficient and reconfigurable low-latency hardware accelerator for spiking neural networks that address these research gaps.  We highlight the contributions of our work below:

    \begin{itemize}
  \item  \emph{High-accuracy SNNs optimized for reduced precision representation and low latency}: Building upon the methods in \cite{li2022quantization,bu2023optimal, shymyrbay2022training}, we demonstrate conversion of software-trained full-precision Resnet-18 deep network to a reduced precision SNN that is amenable for hardware implementation, without appreciable loss in accuracy and requiring $<8$ timesteps.  
  \item \emph{Area and power-efficient processing element for spiking convolutional acceleration}: We demonstrate a reconfigurable spiking inference accelerator (SIA) by designing a novel processing element consisting of 3 multiplexers and an 8-bit adder, ideal for implementing $3\times3$ convolutional kernels, avoiding expensive multipliers. We also demonstrate the reconfigurability of our accelerator by quantifying the performance of this  PE for supporting other kernel sizes and fully-connected layers.  
    \end{itemize}
  
\section{Background}\label{sec: Background}

\noindent \emph{Spiking Neural Networks:} In spiking neural networks,  signals  $\boldsymbol{s_t}$ used for encoding and transmitting data are binary  (`1' and `0'),  and have a duration of $T$ timesteps, where $i^{th}$ component of $\boldsymbol{s_t}$, $s_{i,t}\in\{0, +1\}$.   
Each neuron $i$ maintains an internal state called membrane potential $U_i$ and  spikes as per the condition
\begin{equation}
    s_{i,t} = \Theta(U_i-\theta)
    \label{eq:heaviside}
\end{equation}
with $\Theta(\cdot)$ being the Heaviside step function and $\theta$ is a learnable threshold for the different layers of the network. Upon emission of a spike, the membrane potential resets either to zero or soft resets to a smaller value, referred to as reset-by-subtraction \cite{bu2023optimal}. We use reset-by-subtraction for our work as this approach has demonstrated better classification accuracy. 

\subsection{Hardware-friendly ANN to SNN Conversion}
We use a general framework to convert a  pre-trained ANN to SNN. This involves 3 steps (Fig.~\ref{fig2}) \cite{bu2023optimal}:
\begin{itemize} 
    \item Train an ANN (with FP32 precision) via traditional training methods e.g., back-propagation;
    \item Quantize the network by replacing the ReLU activation in ANN with a quantized ReLU of $L$ levels, and train for step size $s^l$, where $l$ is the given neuronal layer. 
    \item Replace the Quantized ReLU with IF layer with threshold $s^l$. (all parameters in INT8 precision).
\end{itemize}

\ignore{ 
The equation for step size is given by

\begin{equation}
    y^l = s^l \times \bigg\lfloor \frac{1}{L} \frac{v^{l}L}{s^l}, 0, 1 \bigg\rceil
    \label{ANN-SNN}
\end{equation}

where $v^l$ is the pre-activation.}



\section{Proposed Architecture Design Methodology}\label{sec:architecture}
In this section, we discuss the salient features of proposed spiking inference accelerator (SIA) that is implemented with the programming logic (PL).
As shown in Fig. \ref{Fig} the proposed architecture consists of (i) a spiking core, that includes 64 processing elements (PE) to implement event-driven synaptic integration  (ii) an aggregation core to execute batch normalization and activation functions (both LIF and IF models),  (iii) control and configuration logic, and (iv) a memory unit to store spike inputs, network weights, and spike outputs.  The details of each of these blocks are described below.

\subsection{Spiking Core and Processing Elements} 
The spiking core is comprised of  64 processing elements (PE) arranged as an $8\times8$ array. Each processing element has three 8-bit multiplexers and an 8-bit adder. To compute a neuron's membrane potential,  synaptic weights corresponding to its presynaptic neurons should be accumulated when they spike. We consider the spiking convolution operation through a $3\times3$ kernel, the analysis can be extended to other kernel sizes and fully connected layers as explained in other works  \cite{nimbekar2022low,chen2019eyeriss}. We accumulate weights row-by-row in an event-driven fashion,  requiring 3 clock cycles for each of the three rows and 1 final cycle to generate the membrane potential. 
 
 As shown in Fig. \ref{Fig}, each PE has 3 multiplexers and an adder. One of the inputs to the multiplexer is set to zero (0) and the other input is kernel weight data (W1, W2, W3). The incoming input spike data is used to select between weights/zero in the multiplexer.  An 8-bit adder accumulates the three inputs from the multiplexers with the partial sum till all the rows of the kernel are computed. This accumulated partial sum (16 bits) is then transferred to the aggregation core.  

\subsection{Aggregation core} 

The aggregation core consists of a dedicated unit to perform activation (spike generation) and batch normalisation. 

\noindent \emph{Activation}: The accumulated partial sum is added to the membrane potential from the previous time step to determine if it is above the spiking threshold (16-bit). Proposed architecture supports setting unique thresholds for each layer of the network and both LIF and IF computation. 
As described in Fig \ref{Fig} the activation function has 2 modes (LIF and IF) based on. If the mode is low (0) then IF neuron activation is performed. If the mode is high (1), LIF neuron activation is performed. 
After a spike, we implement reset-by-subtraction \cite{rueckauer2017conversion}, wherein the membrane potential is set to the difference between the current membrane potential and threshold voltage. Otherwise, no spike is issued (output is 0), and the updated membrane potential is stored in the input scratchpad memory for the next time step.\\ 
\noindent \emph{Batch Normalisation}: In CNNs, batch normalization or batchnorm operation is typically shifted to preceding layers, and can thus be performed in PEs \cite{perez2021heterogeneous, liu2023efficient}. However, in SNNs batchnorm operates on accumulated spikes and therefore involves real-valued multiplications, performed by fixed-point multipliers, and bias addition performed separately (bias is added to the batch normalisation output). We also observe that we get better accuracy when accumulated spikes and batchnorm coefficients are represented in higher precision (16-bit). Since we do not have multipliers in PEs, we have special provision in aggregation core to perform batch normalization, involving the following operation  

\vspace{-2mm}
\begin{equation}
    y_{bn} = \frac{y-\mu}{\sqrt{(\sigma^2 + \varepsilon)}}\gamma + \beta  \equiv yG + H.
    \label{eq:bn}
\end{equation}
Here, $\mu$ and $\sigma^2$ are the running mean and variance estimates, $\varepsilon$ is an arbitrarily small value for numerical stability, and $\gamma$ and $\beta$ the batchnorm affine terms that are learnt during training. To implement batchnorm in our hardware, we use the transformed variables $G = \gamma q_w/\sqrt{(\sigma^2 + \varepsilon)}$ and $H = \mu G/q_w - \beta$, where $q_w$ is the scaling factor for weight quantization, which is also a learnable parameter. The $G$ and $H$ are transferred from processor to the SIA layerwise as part of the configuration mention in Fig. \ref{Fig}. 
\ignore{
Operations like multiplication and division are resource consuming to perform and therefore need to be reduced. We transforms the values of batchnorm parameters to coefficients $G$ and $H$ such that
\begin{equation}
    y_{bn} = yG + H.
    \label{eq:bn}
\end{equation}
This halves the total number of operations and additionally makes it easier from fixed-point design point of view, since number of quantizations also halve. Here $G = \gamma q_w/\sqrt{(\sigma^2 + \varepsilon)}$ and $H = \mu G/q_w - \beta$. $q_w$ is the scaling factor for weight quantization, which is a learnable parameter.}

\subsection{Control and Configuration}
This includes circuits for the management of the spiking core,  aggregation core, and memory units.  A dedicated controller unit is designed to manage memory access and core computation operations. 

%
\subsection{Memory organisation}
The optimization of memory allocation and data flow plays a crucial role in maximizing throughput and efficiency. 
Compared to ANNs, SNNs require more data transfer operations between the processor (PS) and the programmable logic (PL), as each input pattern is encoded with binary signals lasting $T$ timesteps. Therefore we propose the following memory organization specifically for the spike input and internal state (membrane potentials) data. 
The spike input memory is divided into three parts -- 128 bytes to store incoming spikes, 128 kB to store parameters of the residual layers, and 64 kB to store membrane potentials. Weight memory is used to store the kernel weights of the network -- we provision 8 KB such that up to 64 kernels can be stored. The output memory (56 kB) is used to store the output spiking data. 


\begin{figure}[!h]
\begin{subfigure}{0.49\columnwidth}
\includegraphics[width=\textwidth]{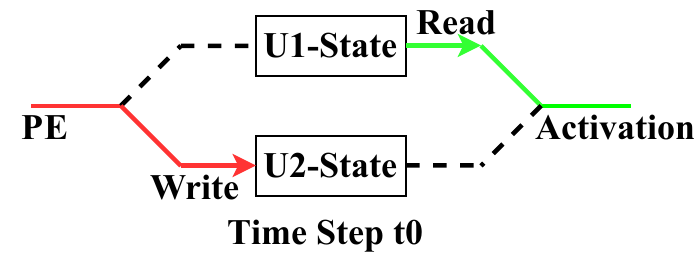}
\caption{}
\label{u1}
\end{subfigure}
\hfill
\begin{subfigure}{0.49\columnwidth}
\includegraphics[width=\textwidth]{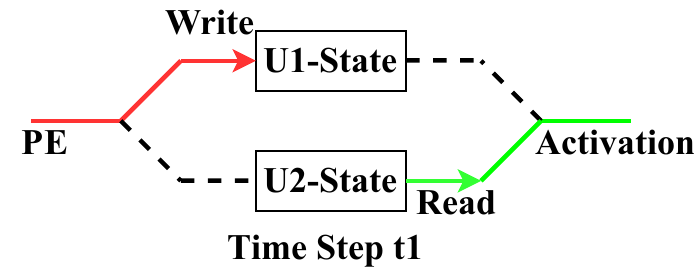}
\caption{} 
\label{u2}
\end{subfigure} 
\caption{Ping-pong protocol for  membrane potential storage (green: read, red: write)}
\label{pingpong}
 \end{figure}
 
The 64 kB allocation to store membrane potentials is configured to operate in a ping-pong fashion \cite{joo1998doubling}. The total available memory is divided into two equal parts named U1-State and U2-State to store and access the membrane potential between the PE and the activation unit. 
As shown in Fig. \ref{pingpong}, at any time step, one part of the memory is used to store the membrane potentials from the PE to the memory, and the other part is used to read the stored membrane potentials value to transmit to the activation unit. At some time step (denoted t1), the membrane potential from the previous time step t0 is added to?;  hence U1-State memory is in read mode while the U2-State memory is in write mode as illustrated in Fig \ref{u1}. Similarly, at time step t2, U1-State memory is in the write mode and U2-State memory is in read mode (Fig \ref{u2}). The toggling of memory is organized in this fashion to read and write the internal states (membrane potential).




 \begin{figure}[!h]
      \centering  \includegraphics[width=0.38\textwidth]{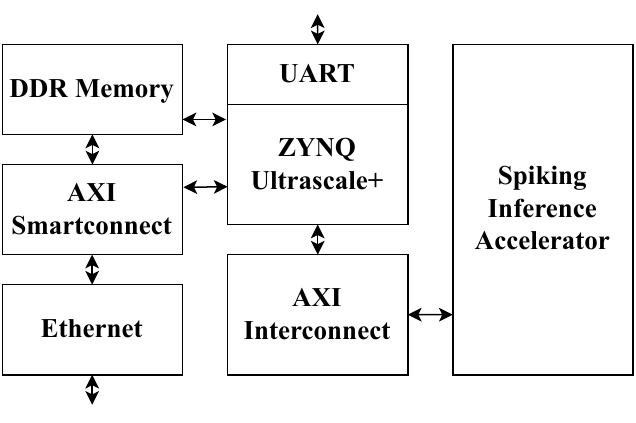} 
    \caption{High-level block diagram of the FPGA Implementation of our spiking inference accelerator (SIA).}
    \label{WBCD_result}
\end{figure} 
 
\ignore{
\begin{figure}[h]
  \centering
    \includegraphics[width=0.5\textwidth]{fig5_v2.png}
    \caption{Ping-pong protocol for membrane potential storage (green: read, red: write)}
    \label{ustate}
\end{figure}
 }


\section{ Hardware Implementation}  
Fig. \ref{WBCD_result} illustrates the FPGA implementation of the proposed SIA architecture. We use a Xilinx PYNQ-Z2 board, equipped with a ZYNQ processor. The synthesis was carried out using Xilinx Vivado 2019.1, operating at a clock frequency of 100 MHz.  
The ZYNQ processor implements the frame data conversion for non-spiking inputs or can transfer event-driven data streams directly to the SIA.  
The DDR memory stores both the parameters of the SNN model and the input data, offering a centralized repository. Data is transferred from an external host to the DDR memory through the ethernet interface. The UART interface is utilized for console access, providing a user-friendly interface for monitoring and troubleshooting. We use AXI4 lite protocol for transferring data between PS and PL.

\begin{figure}[h]
  \centering
    \includegraphics[width=0.45\textwidth]{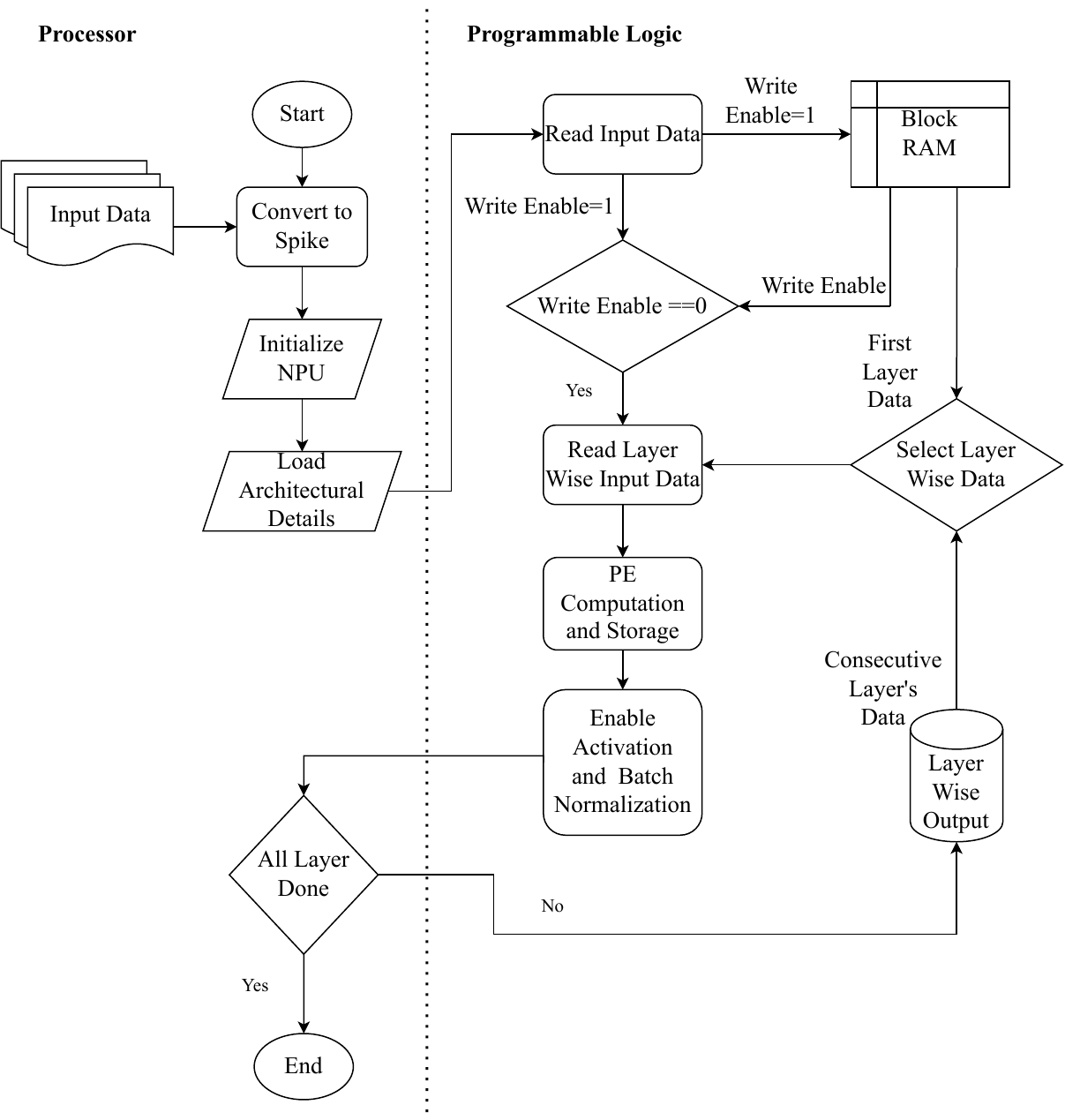}
    \caption{Implementation flow to configure the proposed SIA architecture on FPGA.}
  \label{flow}
\end{figure} 
Fig. \ref{flow} shows the  FPGA implementation flow of the proposed architecture. 
Once the spikes and kernels for a layer are streamed into the Block RAMs in the PL, 
PE executes spiking convolutions to generate the partial sum. This is passed on to the batch normalization unit followed by the spiking activation unit to determine the output spike state. For residual layers,  pre-computed partial sums are read from the processor which is accumulated with the partial sums present in the PL before batch normalization and spiking activation. This process is repeated for all layers of the network sequentially. 

\section{Implementation Results and Analysis}
We first discuss the classification accuracy for CIFAR-10 dataset with the 11M parameter Resnet-18 and VGG-11 network converted to spike domain (Fig. \ref{acc1}, \ref{acc2}).  
For Resnet-18 (Fig. \ref{acc1}) the baseline FP32 ANN accuracy is $95.83\%$ (blue) and after quantisation of the membrane potential and ReLU the accuracy is $94.37\%$ (red). After replacing the quantised ReLU units by IF spike activation model, with the learnt threshold values used for each layer we got the SNN accuracy of $94.71\%$ (black). We report the 8-bit SNN accuracy as a function of  the number of timesteps used for signal encoding and model execution; the SNN accuracy (black points $94.71\%$) exceeds the quantised ANN accuracy after 8 timesteps, and settles to within $<1\%$ of the baseline ANN accuracy.

Similarly for VGG-11, the baseline ANN accuracy is $91.25\%$ (blue) and after quantisation of membrane potential the accuracy is $90.05\%$ (red). However, after replacing ReLU by IF activation we get an accuracy of $90.47\%$ (black) as shown in Fig. \ref{acc2}.
\begin{figure}[h]
  \centering
    \includegraphics[width=0.45\textwidth]{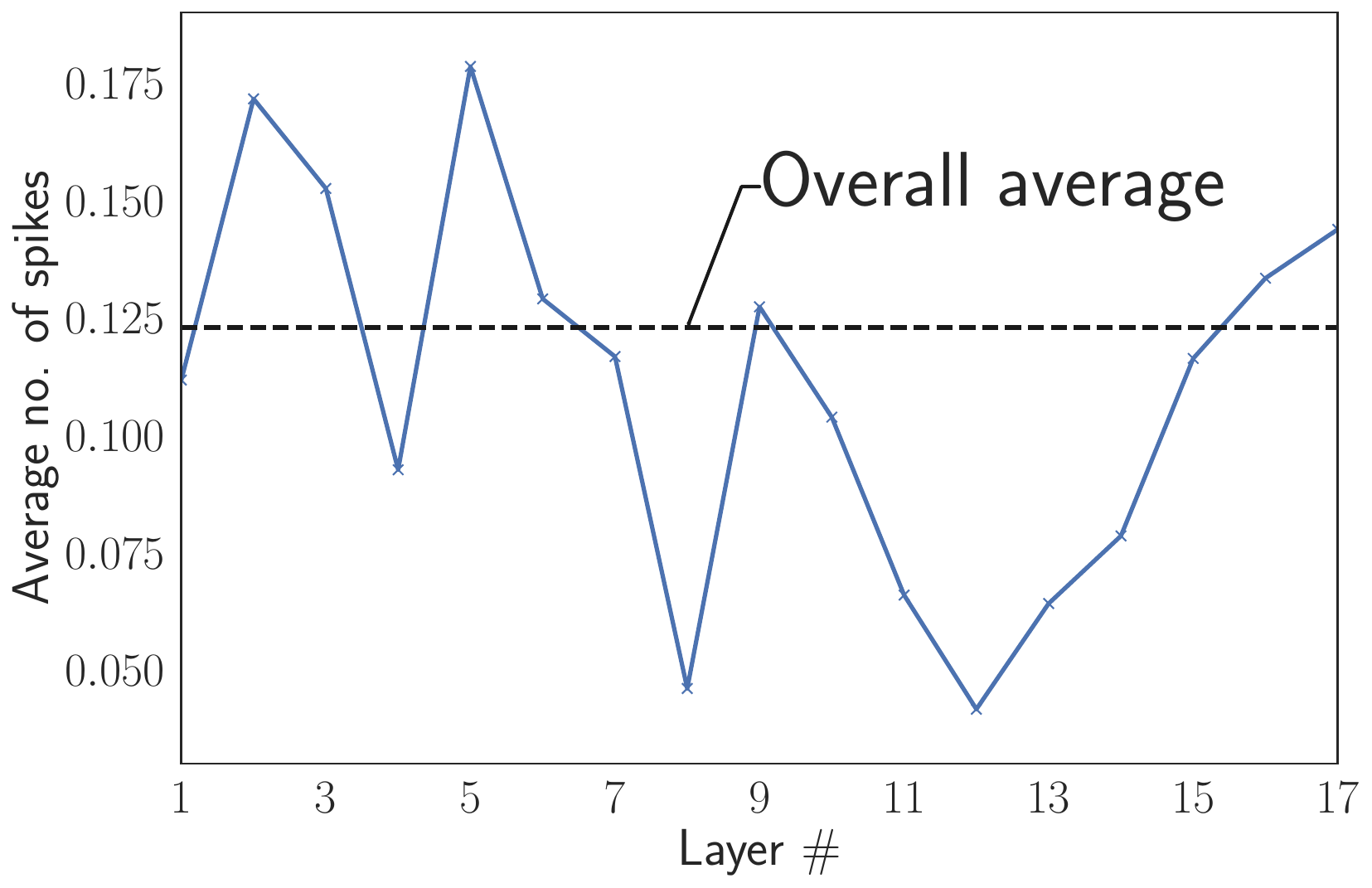}
    \caption{Average spike rate across layers of ResNet-18 of the optimised network}
  \label{avg_spikerate}
\end{figure} 
We also calculate the average spike rate per layer for both the networks. The overall average spike rate was found to be around 0.12 and 0.16 spikes per timestep for ResNet-18 and VGG-11 respectively, and no significant decreasing trend in spikes were seen in deeper layers (Fig. \ref{avg_spikerate}, \ref{avg_spikerate_vgg}). This is expected as we use reset-by-subtraction and adjust threshold at each layer, which results in higher spike rates \cite{rueckauer2017conversion,lu2020exploring}. Thus we trade-off high spike rates for faster inference.


\begin{figure}[]
\centering
\captionsetup{justification=centering,margin=0.35cm}
  \includegraphics [scale = 0.3]{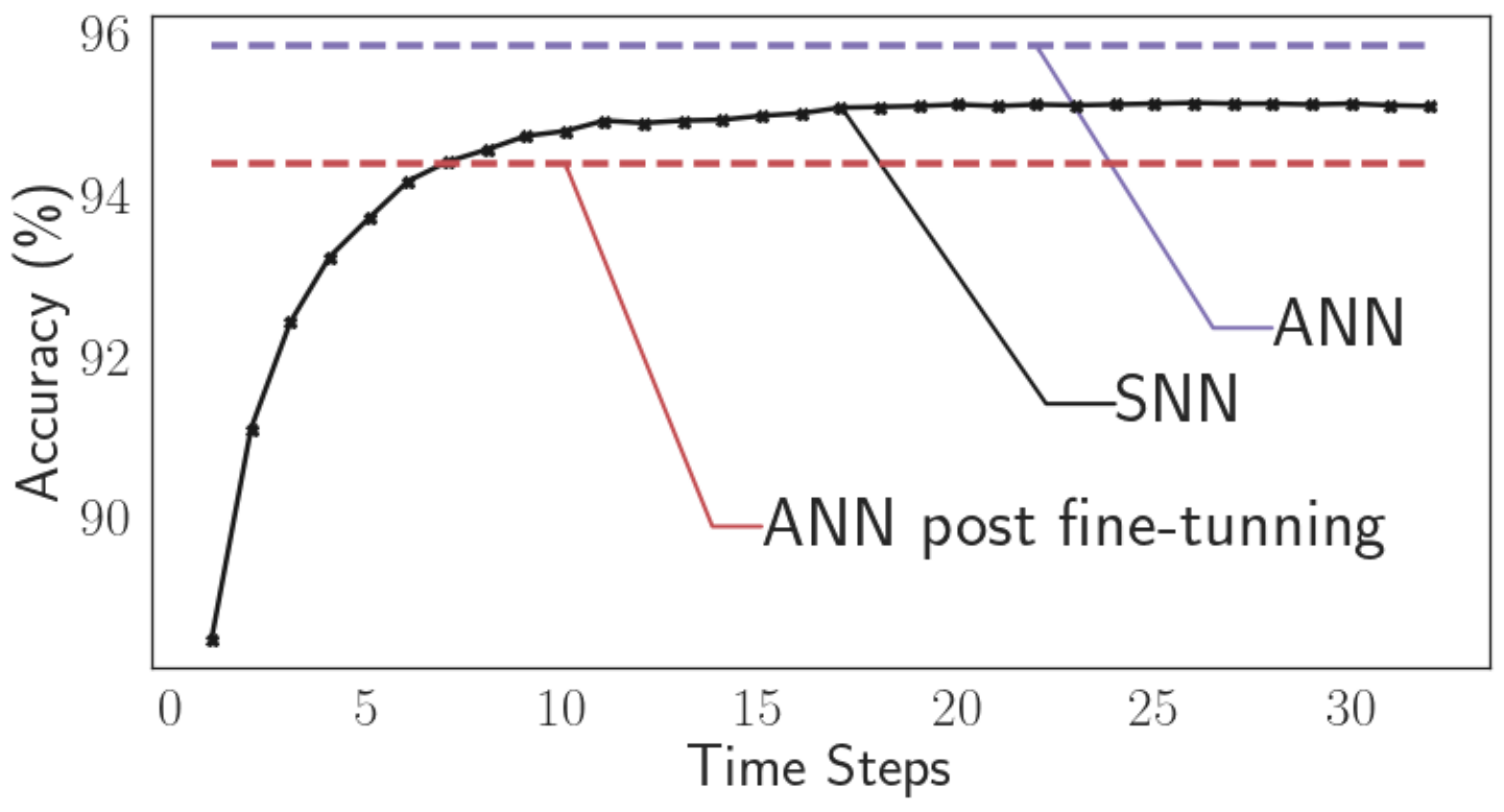}
  \caption{Classification accuracy of 8-bit Resnet-18 SNN for CIFAR-10 dataset, as a function of spike timesteps.}
  \label{acc1}
\end{figure} 
\begin{figure}[]
  \centering
    \includegraphics[width=0.45\textwidth]{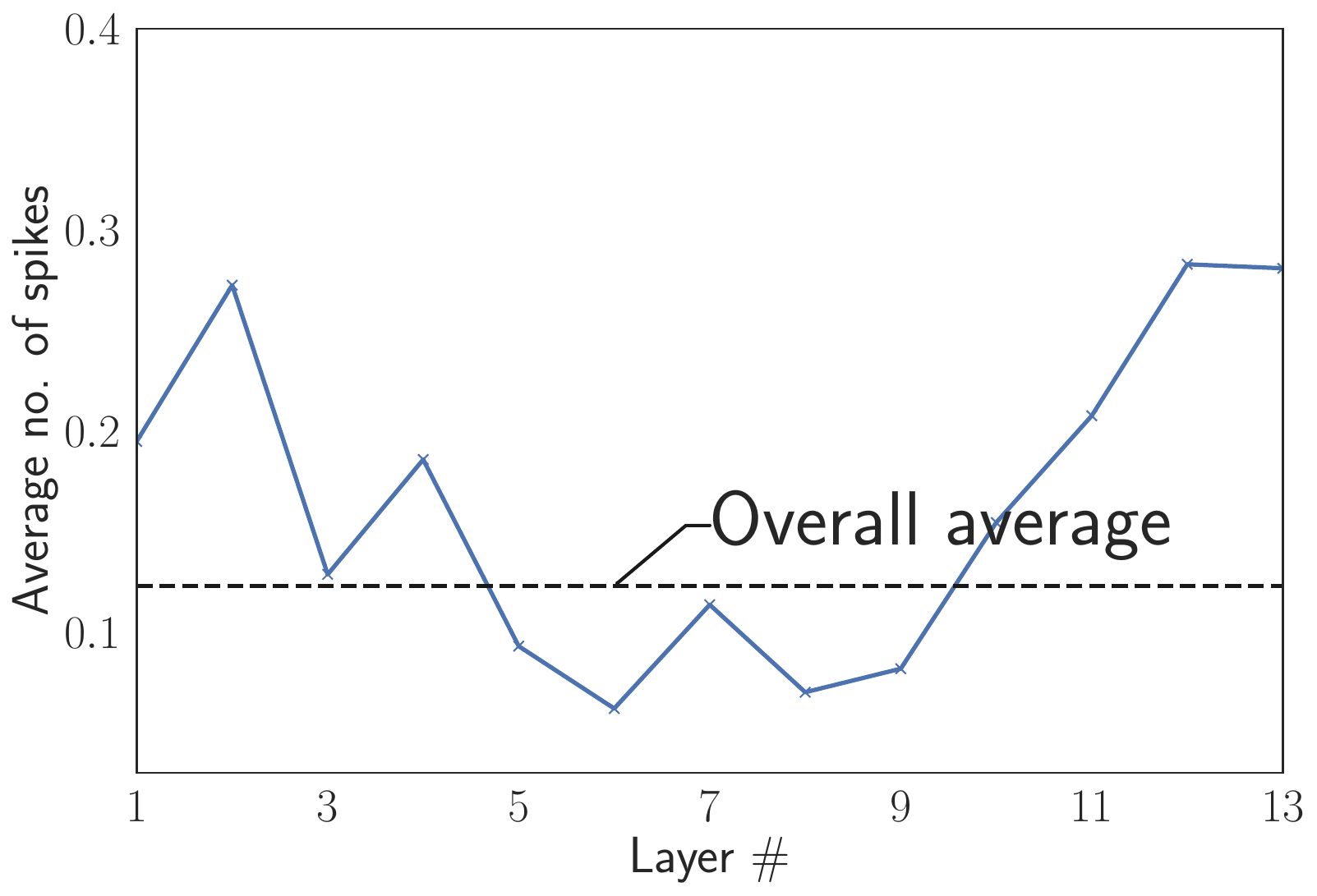}
    \caption{Average spike rate across layers of VGG-11 of the optimised network.}
  \label{avg_spikerate_vgg}
\end{figure}

\begin{figure}[]
\centering
\captionsetup{justification=centering,margin=0.35cm}
  \includegraphics [scale = 0.3]{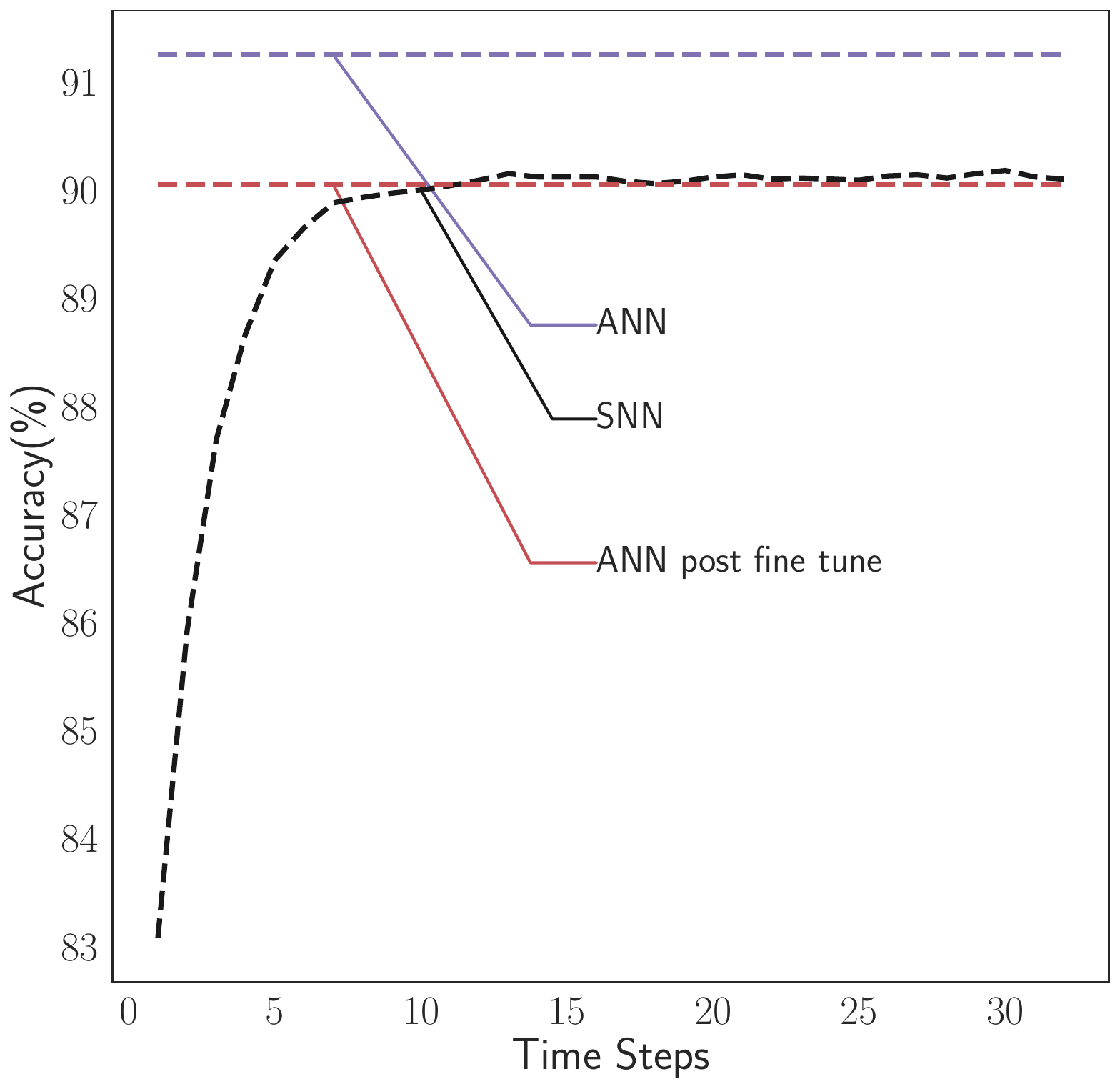}
  \caption{Classification accuracy of 8-bit VGG-11 SNN for CIFAR-10 dataset, as a function of spike timesteps.}
  \label{acc2}
\end{figure} 
 

\begin{table}[H]
\begin{center}
\caption{Layer-wise latency for 8-bit ResNet-18 AND VGG-11.}
\label{tab_latency}
\begin{tabular}{cccc}
\cline{1-3}
\multicolumn{3}{|c|}{ResNet-18}                                                                               &  \\ \cline{1-3}
\multicolumn{1}{|c|}{Layer}            & \multicolumn{1}{c|}{Output size} & \multicolumn{1}{c|}{Latency (ms)} &  \\ \cline{1-3}
\multicolumn{1}{|c|}{Conv 5 (3x3,64)}  & \multicolumn{1}{c|}{32x32}       & \multicolumn{1}{c|}{4.73}         &  \\ \cline{1-3}
\multicolumn{1}{|c|}{Conv 4 (3x3,128)} & \multicolumn{1}{c|}{16x16}       & \multicolumn{1}{c|}{3.58}         &  \\ \cline{1-3}
\multicolumn{1}{|c|}{Conv 4 (3x3,256)} & \multicolumn{1}{c|}{8x8}         & \multicolumn{1}{c|}{3.58}         &  \\ \cline{1-3}
\multicolumn{1}{|c|}{Conv 4 (3x3,512)} & \multicolumn{1}{c|}{4x4}         & \multicolumn{1}{c|}{3.57}         &  \\ \cline{1-3}
\multicolumn{1}{|c|}{FC (512)}         & \multicolumn{1}{c|}{512x10}      & \multicolumn{1}{c|}{58.929}         &  \\ \cline{1-3}
\multicolumn{3}{|c|}{VGG-11}                                                                                  &  \\ \cline{1-3}
\multicolumn{1}{|c|}{Layer}            & \multicolumn{1}{c|}{Output size} & \multicolumn{1}{c|}{Latency (ms)} &  \\ \cline{1-3}
\multicolumn{1}{|c|}{Conv (3x3,64)}    & \multicolumn{1}{c|}{32x32}       & \multicolumn{1}{c|}{0.94}         &  \\ \cline{1-3}
\multicolumn{1}{|c|}{Conv (3x3,128)}   & \multicolumn{1}{c|}{16x16}       & \multicolumn{1}{c|}{0.89}         &  \\ \cline{1-3}
\multicolumn{1}{|c|}{Conv 2 (3x3,256)} & \multicolumn{1}{c|}{8x8}         & \multicolumn{1}{c|}{2.68}         &  \\ \cline{1-3}
\multicolumn{1}{|c|}{Conv 3 (3x3,512)} & \multicolumn{1}{c|}{4x4}         & \multicolumn{1}{c|}{2.67}         &  \\ \cline{1-3}
\multicolumn{1}{|c|}{FC (512)}                              & \multicolumn{1}{c|}{512x10}                          & \multicolumn{1}{c|}{ 58.72}                             &  \\ \cline{1-3}
\end{tabular}
\end{center}
\end{table}

Table \ref{tab_latency} shows the layer-wise computational latency for the Resnet-18 and VGG-11 on the PYNQ-Z2 FPGA; the latency for other kernel dimensions are reported in Table \ref{tab_latency3}. Table \ref{tab2} shows the resource utilization of the proposed architecture on the PYNQ-Z2 FPGA, which consumes $1.54\,$ Watts. 
The PE efficiency and energy efficiency of our design is $600\,$MOPs/PE and $24.93\,$GOPs/W, the highest reported in the literature (Table \ref{fpga_tab}). Our design also uses fewer multipliers resulting in less DSP block utilization  (17) and power consumption.  We also synthesized the SIA architecture with TSMC 40 nm technology projecting a  throughput of 192 GOPS with a frequency of 500 MHz consuming $11\,$mm$^2$ and $2.17\,$W.

\begin{table}[]
\begin{center}
\caption{Latency as a function of  8-bit kernel dimensions.}
\label{tab_latency3}
\begin{tabular}{|c|c|c|}
\hline
Conv (3x3,64) & 32$\times$32 & 0.9479     \\
\hline
Conv (5x5,64) & 32$\times$32 & 0.95      \\
\hline  
Conv (7x7,64)  & 32$\times$32 & 0.9677 \\
\hline
Conv  (11x11,64)  & 32$\times$32&  0.9839     \\    
\hline 
\end{tabular}
\end{center}
\end{table} 

\begin{table}[]
\begin{center}
\caption{FPGA resource utilization.}
\label{tab2}
\begin{tabular}{|c|c|c|c|}
\hline
Parameter & Utilized & Available & Percentage \\
\hline
LUTs & 11932 & 53200  & 22.43\%  \\
\hline
FFs & 8157 & 105400  & 7.67\%  \\
\hline
DSPs & 17 & 220 & 7.67\% \\
\hline
BRAMs & 95 & 140 & 67.86\% \\  \hline
LUTRAMs & 158 & 17400 & 0.90\%  \\ \hline 
BUFG & 1 & 32 & 3.13\%  \\\hline 
\end{tabular}
\end{center}
\end{table}

\begin{table}[]
\centering
\caption{Performance comparison  with prior art.}
\label{fpga_tab}
\begin{tabularx}{\columnwidth}{|p{1.4cm}|X|X|X|X|X|X|}
\hline
Paper & \cite{gilan2019fpga} & \cite{qiu2016going} & \cite{chen2020throughput} & \cite{li2021fpga} & \cite{guo2017angel} &  \textbf{This Work} \\
\hline
Platform & ZC706 & ZC706 & VC707 & VC709 & XC7Z020 &   PYNQ-Z2\\
\hline
\# of PEs  & 576 & 780 & 64 & 664 & 12     &64\\
\hline 
Clock Freq. (MHz) &  200 & 150 & 200 & 200 & 200   &100 \\
\hline 
Throughput (GOPS)  & 198.1 & 187.8 & 12.5 & 220 & 187.80   &38.4 \\
\hline 
PE Eff. (GOPs/PE) & 0.343 & 0.241 & 0.195 & 0.331 & N/A  & 0.6\\
\hline 
Energy Eff. (GOPs/W) & N/A & 14.22 & N/A & 22.9 & 19.50 &24.93 \\
\hline 
DSP & 576 & 780& N/A & 664 & 400 & 17\\
\hline 
GOPS/DSP & 0.34 & 0.24 & N/A & 0.33 & 0.46 & 2.25\\
\hline
 
\end{tabularx}

\end{table}

\section{Conclusion}
This paper demonstrated a hardware-software co-optimization method for accurate and fast SNN-based inference. The hardware accelerator design is demonstrated to be highly resource-efficient,  achieving $0.6\,$GOPS per processing element and $2.25\,$GOPS/DSP slice, corresponding to  $2\times$ and $4.5\times$ higher utilisation efficiency respectively compared to the state-of-the-art. Building upon the experimental FPGA validation achieving  $25\,$ GOPS/W, future work will focus on a $600\,$GOPS/W, $40\,$nm CMOS ASIC design and tape-out,  targeting competitive performance with state-of-the-art \cite{tianjic}.  

\section{Acknowledgement} This research was supported in part by the EPSRC Open Fellowship EP/X011356/1 and the EPSRC grant EP/X011852/1. Anagha Nimbekar and Amit Acharyya would like to acknowledge the support received form the Defence Research Organization (DRDO) and Ministry of Electronics and Information Technology (MEITY), Government of India.
 
 
\bibliography{name}
 \ignore {

}
\end{document}